\documentclass[journal]{IEEEtran}

\ifCLASSINFOpdf
\else
   \usepackage[dvips]{graphicx}
\fi
\usepackage{url}

\usepackage{graphicx}
\usepackage{graphics} 
\usepackage{epsfig} 
\usepackage{mathptmx} 
\usepackage{times} 
\usepackage{amsmath} 
\usepackage{amssymb}  
\usepackage{amsfonts}
\usepackage{booktabs}
\usepackage{xcolor}

\def\etal{\textit{et al.}}

\usepackage{hyperref}

\usepackage{mathrsfs}

\usepackage{textcomp} 
\DeclareSymbolFont{textsymbols}{TS1}{\familydefault}{m}{n}
\SetSymbolFont{textsymbols}{bold}{TS1}{\familydefault}{m}{n}

\DeclareMathSymbol{\ulq}{\mathopen}{textsymbols}{39}
\DeclareMathSymbol{\urq}{\mathclose}{textsymbols}{39}
\DeclareMathAlphabet{\mathcal}{OMS}{cmsy}{m}{n}

\begin{document}

\urlstyle{tt}

\title{\LARGE{Unsupervised Multiview Contrastive Language-Image Joint Learning \\with Pseudo-Labeled Prompts Via Vision-Language Model \\for 3D/4D Facial Expression Recognition}}

\author{Muzammil Behzad, \IEEEmembership{Member, IEEE}

}

\maketitle

\begin{abstract}
In this paper, we introduce MultiviewVLM, a vision-language model designed for unsupervised contrastive multiview representation learning of facial emotions from 3D/4D data. Our architecture integrates pseudo-labels derived from generated textual prompts to guide implicit alignment of emotional semantics. To capture shared information across multi-views, we propose a joint embedding space that aligns multiview representations without requiring explicit supervision. We further enhance the discriminability of our model through a novel multiview contrastive learning strategy that leverages stable positive-negative pair sampling. A gradient-friendly loss function is introduced to promote smoother and more stable convergence, and the model is optimized for distributed training to ensure scalability. Extensive experiments demonstrate that MultiviewVLM outperforms existing state-of-the-art methods and can be easily adapted to various real-world applications with minimal modifications.
\end{abstract}

\begin{IEEEkeywords}
artificial intelligence, facial expression recognition, emotion recognition, vision-language models (VLMs), 3D/4D point-clouds
\end{IEEEkeywords}

\IEEEpeerreviewmaketitle

\section{Introduction}

\IEEEPARstart{V}{ision} language models have \cite{bordes2024introductionvisionlanguagemodeling} shown exemplary performance over the last few years by extending the large language models (LLMs) \cite{minaee2024largelanguagemodelssurvey} towards the visual domain. One important aspect of these vision language models (VLMs) is their capability to leverage large-scale pre-training on extensive datasets, and then enabling fine-tuning for specific downstream tasks or even strongly supporting zero-shot learning \cite{foteinopoulou_emoclip_2024}. Equipped with computational efficiency and scalability, the emergence of advanced multi-modal models, such as the contrastive language-image pre-training (CLIP)~\cite{radford2021learningtransferablevisualmodels}, has demonstrated remarkable breakthroughs by effectively integrating both visual and textual information contexts \cite{zhai2024finetuninglargevisionlanguagemodels} for adapting these models towards several computer vision tasks.
\begin{figure}[t!]
    \centering
	\includegraphics[width=\linewidth]{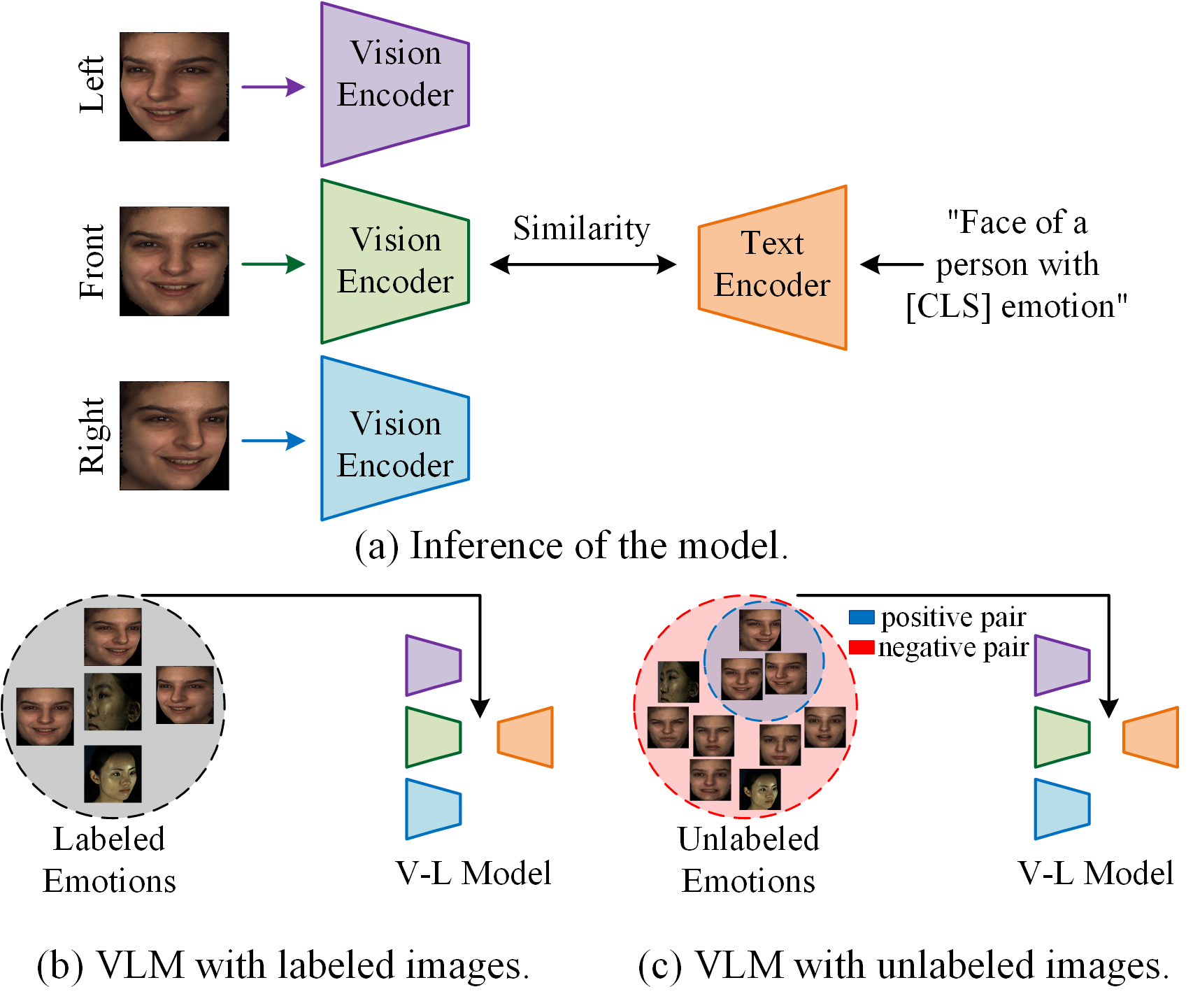}
	\caption{Overview of the vision-language model for emotion recognition: (a) Inference framework leveraging multi-view inputs, (b) Supervised adaptation using labeled target datasets, and (c) Unsupervised joint embedding learning with multi-view positive-negative pairs. We use [CLS] to denote an  emotion class token as: ``Happy", ``Angry", ``Disgust", ``Fear", ``Sad", or ``Surprise".}
	\label{VLmodel}
\end{figure}

As an essential area of research in affective computing, facial expression recognition (FER) focuses on understanding and interpreting human facial emotions, with numerous applications in the fields of human-computer interaction \cite{chowdary2023deep}, mental health \cite{Foteinopoulou_2022}, education \cite{YADEGARIDEHKORDI2019103649}, and beyond \cite{7374704}. In this regard, Ekman \& Friesen's pioneering emotion theory~\cite{ekman1971constants} has laid the foundation for researchers to develop models that address and resolve potential gaps in understanding the complexity of human emotions~\cite{LIU2023423}. Motivated by the scaling success of VLMs, we aim to develop a VLM-based emotion recognition model to uncover and learn from the underlying correlated patterns in 3D/4D FER as shown in Fig.~\ref{VLmodel}.

Various methods have been proposed in the literature to learn from intrinsic 3D facial geometry. In this context, the most commonly used approaches include local feature-based methods~\cite{6460694, 5206613, li2015efficient}, template-based methods~\cite{4539275, 5597896}, curve-based methods~\cite{samir2009intrinsic, maalej2011shape}, and 2D projection-based methods~\cite{7944639, 8265585}. In recent years, 3D/4D FER has attracted significant attention, as it allows deep learning models to capture additional discriminative facial features through the facial depth axis. For instance, Yin~\etal~\cite{4813324} and Sun~\etal~\cite{Sun:2010:TVF:1820799.1820803} utilized Hidden Markov Models (HMM) to extract temporal facial features from 4D facial scans. Similarly, Ben Amor~\etal~\cite{amor20144} highlighted the effectiveness of a deformation vector field based on Riemannian analysis combined with a random forest classifier. Sandbach~\etal~\cite{sandbach2012recognition} also applied HMM and GentleBoost to learn free-form representations of 3D frames. Furthermore, the study in~\cite{FANG2012738} represented geometric coordinates and their normals as feature vectors, while another study~\cite{6130440} employed dynamic local binary patterns (LBP) alongside a support vector machine (SVM) for facial expression recognition. In a related approach, the authors in~\cite{6553746} extracted features from polar angles and curvatures, introducing a spatio-temporal LBP-based feature extractor for recognition.

On the other hand, Li~\etal~\cite{8373807} proposed an innovative framework for automatic 4D Facial Emotion Recognition (FER) using a dynamic geometric image network. This approach involved generating geometric images by computing differential quantities from 3D facial point-clouds. Emotion prediction was performed by fusing probability scores at the score level obtained from multiple geometric images. Building on such advancements, unsupervised learning approaches have enabled the automatic learning of feature representations without manual annotations. Recent methods~\cite{NEURIPS2020_70feb62b, chen2020simple, NEURIPS2020_f3ada80d} have demonstrated that self-learned representations can rival supervised ones by maximizing the similarity of distorted samples and identifying distortion-invariant \cite{zbontar2021barlow} correlated patterns to recognize similar emotions effectively.


\subsection{Motivations}
Although effective, these traditional methods often depend on manually extracted features and localized cues, restricting their performance and adaptability in real-world scenarios. Inspired by the scaling victory of LLMs~\cite{minaee2024largelanguagemodelssurvey} and VLMs~\cite{zhang2024visionlanguagemodelsvisiontasks}, we develop a VLM-based model that uses unsupervised joint visual-text learning to align visual and textual modalities for improved emotion recognition by leveraging facial patterns in 3D/4D faces for better FER. Unlike conventional 2D-based emotion recognition (e.g., \cite{8245803, 8844064, 9226082, 9369001}), this approach classifies facial expressions from 3D/4D faces with spatio-temporal features, with substantial results~\cite{li20113d, zhen2016muscular, 7163090, li2015efficient} confirming its effectiveness. The joint representations from visual and textual data~\cite{zhai2024finetuninglargevisionlanguagemodels} in VLMs improve generalization and robustness across diverse 3D/4D datasets without the need of manual feature engineering. However, the limited size of existing datasets equally presents a further challenge in this research area. 

\subsection{Contributions}
In light of the above discussions, we highlight our contributions below. To our knowledge, the existing literature lacks research on 3D/4D FER models using VLMs in an unsupervised manner, mainly due to the challenges in implementing such models and the complexity of 3D/4D data structures. In this context, we extend CLIP~\cite{radford2021learningtransferablevisualmodels} to introduce MultiviewVLM, a VLM architecture using multiviews in an unsupervised way. The key features of our model are as follows:
\begin{enumerate}
     \item We introduce pseudo-labels for leveraging implicit alignment via generated text prompts as positive pairs, enriching the model's emotional semantic understanding.
    \item We propose a joint embedding space for shared feature learning to integrate multiview visual-textual embeddings in an unsupervised way.
     \item We also propose a novel multiview contrastive loss for positive-negative multiview pair learning with numerical stability to promote the discriminability and generalizability in the model’s representation space.
    \item We formulate a more effective gradient-friendly loss function for smoother model convergence.
    \item We design MultiviewVLM with the capability to leverage distributed training for substantial scalability.
\end{enumerate}
Furthermore, it must be noted that MultiviewVLM is equally applicable to other downstream tasks like, face recognition, anti-spoofing, identity recognition and  several other applications areas where multiviews are used.

\section{MultiviewVLM: Multiview Vision-Language Model}
Our MultiviewVLM model's capability to integrate multiviews offers a significant improvement with its robust and scalable implementation, making it highly effective and ready for deployment solutions with minimal adjustments.
\begin{figure*}[t!]
    \centering
	\includegraphics[width=\linewidth]{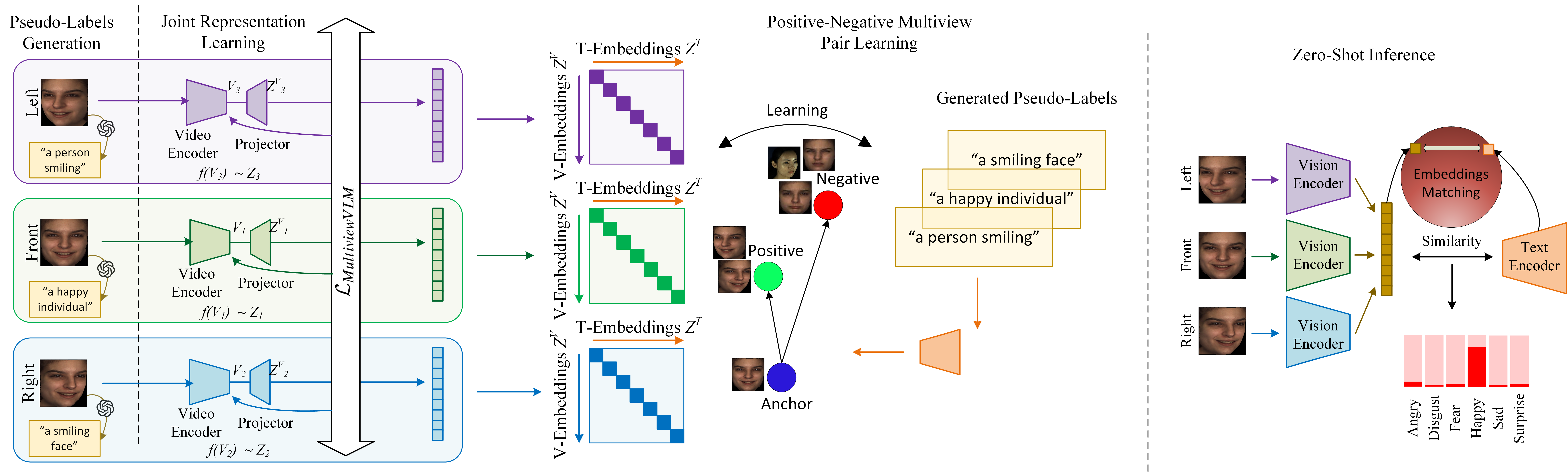}
	\caption{The architecture of MultiviewVLM for unsupervised 3D/4D emotion recognition. The pseudo-labels are generated using a language model such as GPT, and multiview embeddings (front, left, and right) are extracted via vision encoders. The joint representation learning then aligns the textual and visual embeddings in a shared embedding space, while positive-negative pair learning ensures discriminability and consistency across views. The proposed loss $\mathcal{L}_{\text{multiviewVLM}}$ facilitates optimization in the embedding space, resulting in robust emotion representations. During inference, the multiview images are fed through the vision encoder to extract and aggregate embeddings, which are then matched against precomputed textual embeddings using similarity, and then assigning the most semantically relevant emotion label.}
	\label{model}
\end{figure*}

\subsection{Pseudo-labels Generation for Implicit Embeddings Alignment}
To formulate unsupervised learning in vision-language models (VLMs) for 3D/4D facial expression recognition (FER), we introduce a novel pseudo-label generation mechanism that facilitates implicit alignment between visual and textual modalities. Particularly, we initially generate textual prompts as pseudo-labels, denoted by \( \mathcal{P}_{\mathrm{pseudo\text{-}labels}} \), for all the visual data of emotions, ultimately enabling a semantically rich and contextually relevant training pipeline. These pseudo-labels are designed to capture diverse emotional semantics and align multiview representations in the learned embedding space, which helps learning robust representations across varying facial expressions and views.

The textual pseudo-labels are generated using pre-trained language models, such as GPT~\cite{brown2020languagemodelsfewshotlearners}, to create effectively diverse emotion-related textual prompts based on generic descriptions of facial expressions. For example, textual prompts such as ``a smiling face", ``a happy individual", or ``a person smiling" are generated to capture subtle variations of the same emotion ``Happy". These prompts serve as semantic anchors for aligning the embeddings of multiview images (e.g., front, left, and right profiles). By associating these textual embeddings with the corresponding visual embeddings through the proposed framework, we ensure that multiview representations of the same 3D emotions are jointly represented in the shared embedding space. This approach not only demonstrates the model’s emotional semantic understanding but also facilitates robust alignment across both visual and textual modalities, making it particularly effective for handling the complexity of 3D/4D FER data.


\subsection{Joint Embedding Space Alignment}
In our proposed unsupervised MultiviewVLM model, we incorporate a joint embedding space alignment as a critical step towards its success. This shared space is designed to represent both visual and textual modalities in a unified representation, offering consistent and robust alignment across different views and similar semantic descriptions. As shown in Fig.~\ref{model}, the visual embeddings are extracted from the multiview images of facial expressions using multiview visual encoder, while the textual embeddings are generated using a text encoder fed with the pseudo-labels generated for implicit embeddings alignment. Both of these embeddings are then projected into the shared embedding space, where the alignment process ensures that the representations of semantically related inputs becomes close to one another.

To enforce this alignment, we further propose a novel gradient-friendly loss function that optimizes the similarity between positive pairs, i.e., embeddings of images and text representing the same emotion or multiple views of the same subject, while maximizing the separation between negative pairs, i.e., the embeddings of unrelated images and prompts. This alignment is further refined by aggregating embeddings from multiple views of the same subject during inference, promoting cross-view consistency and enabling the model to capture a holistic representation of facial emotions. By jointly optimizing visual and textual embeddings, our proposed framework ensures that the model learns robust, semantically rich, and contextually aware facial representations, effectively yielding a robust emotion recognition pipeline.

\subsection{Multiview Contrastive Loss for Joint Embedding Alignment}
Contrastive loss plays a fundamental role in aligning visual and textual embeddings in the shared embedding space for unsupervised multiview FER. The objective is to ensure that the positive pairs, such as embeddings of multiview images of the same emotion face are brought closer along with their corresponding textual descriptions, while the negative pairs, such as embeddings of unrelated images and prompts, are pushed apart. This alignment process enables the model to learn robust and discriminative representations of emotions across the available multi-modalities. In this regard, the positive pairs consist of embeddings that represent semantically similar entities. These include multiple views of the same facial expressions, e.g., frontal, left, and right perspectives, embeddings of an image and its corresponding textual prompt, e.g., ``a smiling person", and augmented versions of the same image (e.g., classically flipped or cropped instances). Conversely, the negative pairs consist of embeddings representing semantically dissimilar entities, such as images of different emotions (e.g., a happy face vs. an angry face) or mismatched textual prompts and images, e.g., a textual description ``an angry person" paired with an image of ``a smiling face".

Mathematically, let \(v_i\) represent the visual embedding of the \(i\)-th image, \(t_i\) represent the textual embedding of the \(i\)-th prompt, \(N\) represent the batch size, \(\text{sim}(a, b)\) represent the cosine similarity between embeddings \(a\) and \(b\), and \(\tau\) represent the temperature parameter to control the sharpness of similarity scores. The contrastive loss for image-to-text pair is expressed as
\begin{equation}
\mathcal{L}_{\text{image-to-text}} = - \frac{1}{N} \sum_{i=1}^N \log \frac{\exp(\text{sim}(v_i, t_i) / \tau)}{\sum_{j=1}^N \exp(\text{sim}(v_i, t_j) / \tau)},
\end{equation}
which ensures that the embedding \(v_i\) (visual) is aligned with its corresponding \(t_i\) (textual). Similarly, the text-to-image contrastive loss is given by
\begin{equation}
\mathcal{L}_{\text{text-to-image}} = - \frac{1}{N} \sum_{i=1}^N \log \frac{\exp(\text{sim}(t_i, v_i) / \tau)}{\sum_{j=1}^N \exp(\text{sim}(t_i, v_j) / \tau)}.
\end{equation}
This ensures that the embedding \(t_i\) (textual) is aligned with its corresponding \(v_i\) (visual) embedding. The combined contrastive loss is then defined as
\begin{equation}
\mathcal{L}_{\text{contrastive}} = \frac{1}{2} (\mathcal{L}_{\text{image-to-text}} + \mathcal{L}_{\text{text-to-image}}).
\end{equation}

In addition to aligning positive pairs, the optimization also focuses on separating embeddings for negative pairs. For positive pairs, the goal is to minimize the cosine distance, expressed as \(\min \|v_i - t_i\|\). For negative pairs, the objective is to maximize the cosine distance, expressed as \(\max \|v_i - t_j\|\), for all \(j \neq i\). Furthermore, multiview consistency is enforced by aggregating embeddings from multiple views and ensuring alignment with the corresponding text embedding, which can be expressed as
\begin{equation}
\mathcal{L}_{\text{multiviewVLM}} = \frac{1}{N} \sum_{i=1}^N \| \frac{v_{if} + v_{ir} + v_{il}}{3} - t_i \|,
\end{equation}
where $v_{if}$, $v_{ir}$, and $v_{il}$ refer to front, right, and left views, respectively. The proposed contrastive-style loss provides several advantages. First, it ensures semantic alignment by bringing related visual and textual embeddings closer in the shared embedding space. Second, it enhances the discriminative ability of the model by separating embeddings of different emotions. Third, the incorporation of multiview data promotes generalization by ensuring that the model learns consistent representations across varying perspectives and augmentations. Overall, the contrastive loss is instrumental in driving the alignment of visual and textual modalities, enabling the model to learn robust, semantically meaningful, and contextually aware representations of emotions.

\subsection{Positive-Negative Multiview Pair Learning with Numerical Stability}
We also use a positive-negative pair learning approach, denoted by \( \mathcal{A}_{\mathrm{pos+neg-}} \), that enables effective representation learning without the need for labeled data. In this regard, the proposed framework leverages the multiview nature of the dataset to represent more meaningful positive and negative pairs, to ensure robust and discriminative learning in the shared visual-textual embedding space. With ${v}_{i1}, {v}_{i2}, {v}_{i3}$ representing the embeddings of the front, left, and right views of the same subject with emotion $e_i$, the positive pairs are constructed by associating embeddings from these multiviews:
\begin{equation}
\mathcal{P} = \{({v}_{if}, {v}_{ir}), ({v}_{ir}, {v}_{il}), ({v}_{if}, {v}_{il}) \;|\; \forall i \}.
\end{equation}
To ensure consistency across multiview representations, the similarity between embeddings in positive pairs is maximized as:
\begin{equation}
\mathcal{L}_{\text{pos}} = - \sum_{(v_a, v_b) \in \mathcal{P}} \log \sigma(\text{sim}(v_a, v_b)),
\end{equation}
where $\text{sim}(v_a, v_b) = \frac{v_a \cdot v_b}{\|v_a\| \|v_b\|}$ is the cosine similarity, and $\sigma$ is a simple yet effective sigmoid function ensuring bounded gradients for numerical stability.

On the other hand, the negative pairs involve embeddings from different emotions. We let ${v}_{ik}$ represent the embedding of a view $k$ of subject $i$ with emotion $e_i$, and ${v}_{jl}$ represent the embedding of view $l$ of subject $j$ with emotion $e_j$, where $e_i \neq e_j$. The set of the negative pairs can then be defined as:
\begin{equation}
\mathcal{N} = \{({v}_{ik}, {v}_{jl}) \;|\; e_i \neq e_j, \forall k, l \}.
\end{equation}
To promote discriminability among negative pairs, the similarity between embeddings in negative pairs is minimized as:
\begin{equation}
\mathcal{L}_{\text{neg}} = \sum_{(v_a, v_b) \in \mathcal{N}} \log (1 - \sigma(\text{sim}(v_a, v_b))).
\end{equation}

This approach not only ensures that embeddings from multiviews of the same emotion are consistent and aligned but also enforces significant separation between embeddings of different emotions, enabling the model to capture underlying emotional variations. By leveraging the inherent muscle movements of the facial features with optimized and robust representations in a shared embedding space, our method aligns visual features with the semantic counterparts in an unsupervised manner. This joint optimization substantially improves the model’s robustness, making it capable of recognizing emotions effectively without explicit labels.

\subsection{Gradient Analysis of the Proposed Loss}
Our proposed multiview contrastive loss is inherently a gradient-friendly loss function because its formulation ensures stable and smooth updates to the network parameters during training. To demonstrate this, we derive the gradients of the loss function with respect to the visual and textual embeddings. Subsequently, the image-to-text loss is given as:
\begin{equation}
\mathcal{L}_{\text{image-to-text}} = - \frac{1}{N} \sum_{i=1}^N \log \frac{\exp(\text{sim}(v_i, t_i) / \tau)}{\sum_{j=1}^N \exp(\text{sim}(v_i, t_j) / \tau)},
\end{equation}
where \(\text{sim}(v_i, t_j) = \frac{v_i \cdot t_j}{\|v_i\| \|t_j\|}\) refers to the cosine similarity. For simplicity, let:
\[
s_{ij} = \frac{\text{sim}(v_i, t_j)}{\tau},
\]
and reformulate the loss function as:
\begin{equation}
\mathcal{L}_{\text{image-to-text}} = - \frac{1}{N} \sum_{i=1}^N \log \frac{\exp(s_{ii})}{\sum_{j=1}^N \exp(s_{ij})}.
\end{equation}

The gradient of \(\mathcal{L}_{\text{image-to-text}}\) with respect to the visual embedding \(v_i\) is:
\begin{equation}
\frac{\partial \mathcal{L}_{\text{image-to-text}}}{\partial v_i} = - \frac{1}{N} \sum_{i=1}^N \frac{\partial}{\partial v_i} \log \frac{\exp(s_{ii})}{\sum_{j=1}^N \exp(s_{ij})}.
\end{equation}

Using the chain rule, we compute:
\[
\frac{\partial \log x}{\partial x} = \frac{1}{x}, \quad \frac{\partial s_{ij}}{\partial v_i} = \frac{t_j}{\tau \|v_i\|} - \frac{(v_i \cdot t_j) v_i}{\tau \|v_i\|^3}.
\]
Substituting these, we have:
\begin{equation}
\frac{\partial \mathcal{L}_{\text{image-to-text}}}{\partial v_i} = - \frac{1}{N} \sum_{i=1}^N \left( \frac{\partial s_{ii}}{\partial v_i} - \frac{\sum_{j=1}^N \exp(s_{ij}) \frac{\partial s_{ij}}{\partial v_i}}{\sum_{j=1}^N \exp(s_{ij})} \right).
\end{equation}

In a similar fashion, the text-to-image loss is given as:
\begin{equation}
\mathcal{L}_{\text{text-to-image}} = - \frac{1}{N} \sum_{i=1}^N \log \frac{\exp(s_{ii})}{\sum_{j=1}^N \exp(s_{ji})}.
\end{equation}
The gradient with respect to the textual embedding \(t_i\) is:
\begin{equation}
\frac{\partial \mathcal{L}_{\text{text-to-image}}}{\partial t_i} = - \frac{1}{N} \sum_{i=1}^N \left( \frac{\partial s_{ii}}{\partial t_i} - \frac{\sum_{j=1}^N \exp(s_{ji}) \frac{\partial s_{ji}}{\partial t_i}}{\sum_{j=1}^N \exp(s_{ji})} \right).
\end{equation}

The derivatives of \(s_{ij}\) with respect to \(t_i\) are computed similarly:
\[
\frac{\partial s_{ij}}{\partial t_i} = \frac{v_j}{\tau \|t_i\|} - \frac{(v_j \cdot t_i) t_i}{\tau \|t_i\|^3}.
\]

With the combined contrastive loss is given as:
\begin{equation}
\mathcal{L}_{\text{contrastive}} = \frac{1}{2} (\mathcal{L}_{\text{image-to-text}} + \mathcal{L}_{\text{text-to-image}}),
\end{equation}
its gradient is the average of the gradients of the individual losses. The temperature parameter \(\tau\) ensures stable computation of the gradients, preventing aggressively sharp updates. Moreover, the cosine similarity normalization and the logarithmic smoothing in the loss function further contribute to the excessive stability of gradient updates. These properties make our proposed loss function gradient-friendly, allowing efficient optimization during the alignment of visual and textual embeddings in the shared embedding space.

\subsection{Clustering for Pseudo-Labels with Multiview Aggregation}
To infer emotion categories from visual and textual embeddings in the shared embedding space, we perform clustering as a vital step in our proposed framework. We group similar embeddings into joint clusters, and assign pseudo-labels to incoming samples, ultimately enabling our model to iteratively improve its representation of emotion categories. Afterwards, we perform multiview aggregation that further enhances the clustering process by leveraging multiple perspectives of the same subject, ensuring that the embeddings capture a holistic and robust representation of emotions without being overfitted to pre-learned representations during positive-negative pair learning. In practice, the pseudo-labels generated from clustering are used to guide the model's learning. Over iterative training rounds, the embedding representation become more discriminative, allowing for improved clustering in subsequent rounds. More importantly, the multiview aggregation significantly improves the clustering process by enhancing robustness, as aggregating features across views reduces noise and focuses more on robust representations, leading to more reliable pseudo-labels. Moreover, the multiview embeddings capture diverse spatial details, improving the model’s understanding of expressions. Such learning consistency ensures that the model generalizes well to varying views and conditions, enhancing the performance on unseen data.

\subsection{Performance Scalability with Distributed Leaning}
One of the key design features of our model is its support for distributed training which significantly improved the performance by leveraging available computational resources. In this regard, we utilize PyTorch's \url{torch.distributed} package in combination with NVIDIA's NCCL backend to efficiently distrubute training activities across devices. The implementation automatically determines the optimal configurations, such as, multi-GPU, single-GPU, or CPU, based on current resource availability. When multiple GPUs are available, we employ the \url{ddp} strategy from \url{pytorch-lightning} to enable synchronized training paradigms. In order to prevent communication bottlenecks during training, our system dynamically assigns network \url{IP} addresses and \url{port} values, establishing flexible TCP connections for multi-process coordination. This approach ensures scalable and resource-efficient training of the MultiviewVLM model across varied computing environments. This setup not only ensures resource-efficient training of the VLM-based models but also enables seamless scaling across diverse compute environments.


\section{Results and Discussions}
We evaluate and validate the performance of our proposed MultiviewVLM model using four widely recognized benchmark datasets: Bosphorus~\cite{savran2008bosphorus}, BU-3DFE~\cite{yin20063d}, BU-4DFE~\cite{4813324}, and BP4D-Spontaneous~\cite{ZHANG2014692}. These datasets deliver a diverse range of facial expressions and subject variations, offering comprehensive coverage of both posed and spontaneous affective behaviors in 3D and 4D point-clouds. Following the standard evaluation protocols established in prior works~\cite{8373807, 8023848, 9320291, behzad2021Sparse3D, behzad2021disentangling}, we generate multiview 2D projections from 3D/4D point-clouds to better simulate real-world camera perspectives. For dynamic 4D sequences, we further apply rank pooling~\cite{bilen2018action} to convert the temporal information into a compact dynamic image representation, enabling the model to effectively capture motion and temporal patterns. To the best of our knowledge, there is only one existing \color{magenta}unsupervised method \color{black}for 3D/4D facial expression recognition \cite{behzad2021self}. We include this method in our evaluation for completeness. Additionally, we compare our approach against the state-of-the-art \color{teal}supervised methods\color{black}. Unsupervised baselines are also considered for a comprehensive analysis. To ensure robustness and generalizability, we adopt a 10-fold subject-independent cross-validation strategy across all experiments, which guarantees that no subjects are shared between training and testing splits, thus providing a rigorous evaluation framework for performance benchmarking.

\subsection{Performance on 3D FER}
Following the existing evaluation protocols in prior works \cite{7944639, 8265585}, we conduct experiments on the BU-3DFE and Bosphorus datasets to assess the effectiveness of our proposed unsupervised model for 3D FER. The BU-3DFE dataset consists of 101 human subjects and it is divided into two subsets: Subset I, which includes expressions at two higher intensity levels and serves as the standard benchmark; and Subset II, which is less commonly used in 3D FER and contains expressions at all four intensity levels, excluding the 100 neutral samples. Similarly, following the common practice for the Bosphorus dataset, only the 65 subjects who performed all six basic expressions are considered for evaluation.

\begin{table}[b!]
	\caption{Accuracy ($\%$) comparisons with state-of-the-art methods on the BU-3DFE Subset I and Subset II, and Bosphorus datasets.}
	\label{table:3DFERresults}
	\resizebox{\linewidth}{!}{%
		\begin{tabular}{l c}
			\hline
			Method & Subset I  (\color{blue}$\uparrow$\color{red}$\downarrow$\color{black})\\
			\hline			
			{\color{teal}Zhen \etal \cite{zhen2016muscular}} & 84.50 (\color{blue}4.79$\uparrow$\color{black}) \\ 
			{\color{teal}Yang \etal \cite{7163090}} & 84.80 (\color{blue}4.49$\uparrow$\color{black}) \\
			{\color{teal}Li \etal \cite{li2015efficient}} & 86.32 (\color{blue}2.97$\uparrow$\color{black}) \\
			{\color{teal}Li \etal \cite{7944639}} & 86.86 (\color{blue}2.43$\uparrow$\color{black}) \\ 
			{\color{teal}Oyedotun \etal \cite{8265585}} & 89.31  (\color{red}0.02$\downarrow$\color{black}) \\ 
			\hline
			{\color{magenta}MiFaR \cite{behzad2021self}} & 88.53 (\color{blue}0.76$\uparrow$\color{black})\\ 
			\textbf{{\color{magenta}MultiviewVLM (Ours)}} & \textbf{89.29}\\ 
			\hline
		\end{tabular} 
		\begin{tabular}{l c c}
			\hline
			Method & Subset II (\color{blue}$\uparrow$\color{red}$\downarrow$\color{black}) & Bosphorus  (\color{blue}$\uparrow$\color{red}$\downarrow$\color{black})\\
			\hline
			{\color{teal}Li \etal \cite{li2015efficient}} & 80.42 (\color{blue}3.56$\uparrow$) & 79.72 (\color{blue}0.14$\uparrow$\color{black})\\ 
			{\color{teal}Yang \etal \cite{7163090}} & 80.46 (\color{blue}3.52$\uparrow$) & 77.50 (\color{blue}2.36$\uparrow$\color{black})\\ 
			{\color{teal}Li \etal \cite{7944639}} & 81.33 (\color{blue}2.65$\uparrow$) & 80.00 (\color{red}0.14$\downarrow$\color{black})\\ 
			\hline
			{\color{magenta}MiFaR \cite{behzad2021self}} & 82.67 (\color{blue}1.31$\uparrow$\color{black}) & 78.84 (\color{blue}1.02$\uparrow$\color{black}) \\ 
			\textbf{{\color{magenta}MultiviewVLM (Ours)}} & \textbf{83.98} & \textbf{79.86} \\ 
			\hline
		\end{tabular}
        }
\end{table}
In Table \ref{table:3DFERresults}, our model demonstrates competitive performance on several benchmarks. On Subset I of the BU-3DFE dataset, our proposed MultiviewVLM  model achieves an accuracy of 89.29\%, closely approaching the best-performing supervised model \cite{li2015efficient} which achieves 89.31\%, with only a marginal difference of \color{red}0.02\%\color{black}. Importantly, this performance notably exceeds the accuracy of the competing unsupervised method MiFaR \cite{behzad2021self} by a significant margin of \color{blue}0.76\% \color{black} (89.29\% vs. 85.33\%), underscoring the strength of our representation learning framework.

Similarly, our model achieves a new state-of-the-art result of 83.98\% on Subset II, outperforming all existing methods, both supervised and unsupervised. This is important because the Subset II introduces additional complexity due to the inclusion of more varied intensity levels. More specifically, our model outperforms MiFaR \cite{behzad2021self} and the supervised model~\cite{7944639} by a significant margin of \color{blue}1.31\% \color{black} and  \color{blue}2.65\%\color{black}, respectively, highlighting the robustness of our model to various expression intensities. On the Bosphorus dataset, MultiviewVLM again performs strongly with an accuracy of 79.86\%, closely competing with the top supervised models and surpassing the prior unsupervised approach. These comprehensive comparisons illustrate that despite relying on the unsupervised learning strategy, our model is capable of capturing expressive facial dynamics with a precision that is on par with or superior to the current state-of-the-art supervised models.

\begin{table}[t!]
	\caption{Performance ($\%$) comparison of 4D FER with the state-of-the-art methods on the BU-4DFE dataset.}
	\label{table:4DFERresults}
	\begin{center}
            \resizebox{\columnwidth}{!}{ 
		\begin{tabular}{l c c}
			\hline
			Method & Experimental Settings & Accuracy (\color{blue}$\uparrow$\color{red}$\downarrow$\color{black}) \\
			\hline
			{\color{teal}Sandbach \etal \cite{sandbach2012recognition}} & 6-CV, Sliding window & 64.60 ({\color{blue}31.87$\uparrow$})\\ 
			{\color{teal}Fang \etal \cite{6130440}} & 10-CV, Full sequence & 75.82 ({\color{blue}20.65$\uparrow$})\\ 
			{\color{teal}Xue \etal \cite{7045888}} & 10-CV, Full sequence & 78.80 ({\color{blue}17.67$\uparrow$})\\ 
			{\color{teal}Sun \etal \cite{Sun:2010:TVF:1820799.1820803}} & 10-CV, - & 83.70 ({\color{blue}12.77$\uparrow$})\\
			{\color{teal}Zhen \etal \cite{7457243}} & 10-CV, Full sequence & 87.06 ({\color{blue}9.41$\uparrow$})\\ 
			{\color{teal}Yao \etal \cite{10.1145/3131345}} & 10-CV, Key-frame & 87.61 ({\color{blue}8.86$\uparrow$})\\ 
			{\color{teal}Fang \etal \cite{FANG2012738}} & 10-CV, - & 91.00 ({\color{blue}5.47$\uparrow$})\\ 
			{\color{teal}Li \etal \cite{8373807}} & 10-CV, Full sequence & 92.22 ({\color{blue}4.25$\uparrow$})\\ 
			{\color{teal}Ben Amor \etal \cite{amor20144}} & 10-CV, Full sequence & 93.21 ({\color{blue}3.26$\uparrow$})\\ 
			{\color{teal}Zhen \etal \cite{8023848}} & 10-CV, Full sequence & 94.18 ({\color{blue}2.29$\uparrow$})\\ 
			{\color{teal}Bejaoui \etal \cite{Bejaoui2019}} & 10-CV, Full sequence & 94.20 ({\color{blue}2.27$\uparrow$})\\ 
			{\color{teal}Zhen \etal \cite{8023848}} & 10-CV, Key-frame & 95.13 ({\color{blue}1.34$\uparrow$})\\ 
			{\color{teal}Behzad \etal \cite{behzad2019automatic}}  & 10-CV, Full sequence & 96.50 ({\color{red}0.03$\downarrow$})\\ 
			\hline
			{\color{magenta}MiFaR \cite{behzad2021self}} & 10-CV, Full sequence & 95.76 ({\color{blue}0.71$\uparrow$})\\ 
			\textbf{{\color{magenta}MultiviewVLM (Ours)}} & 10-CV, Full sequence & \textbf{96.47}\\ 
			\hline
		\end{tabular}
             } 
	\end{center}
\end{table}
\subsection{Performance on 4D FER}
To evaluate the effectiveness of our proposed model on 4D FER, we conducted comprehensive experiments on the BU-4DFE dataset, which consists of 3D video sequences of 101 subjects performing six posed facial expressions. Table~\ref{table:4DFERresults} presents the performance comparison with the state-of-the-art methods under similar experimental settings. Our MultiviewVLM model achieves the highest accuracy of 96.47\%, outperforming nearly all existing supervised and unsupervised methods thanks to our joint multiview learning framework. In particular, our model surpasses traditional supervised method~\cite{8023848} by a margin of \color{blue}1.34\% \color{black} and demonstrates stronger generalization compared to methods using key-frame or sliding window strategies. The consistent improvement highlights the strength of our multiview integration and the effectiveness of our loss formulation in capturing spatiotemporal dynamics of underlying 4D expressions. 

Additionally, when compared to the only existing unsupervised baseline, MiFaR \cite{behzad2021self} (95.76\%), our model achieves a clear gain of \color{blue}0.71\%\color{black}, establishing a new benchmark for unsupervised 4D FER. While prior supervised models required extensive annotated training data, our approach reaches comparable or superior performance without the need for manual labels. This not only reduces the annotation burden but also opens new directions for scalable deployment in real world applications. The competitive edge of our MultiviewVLM model with its performance boost over both supervised and unsupervised baselines demonstrates the viability of the joint multiview learning in effectively understanding expressive facial behavior over time.

\begin{table}[b!]
	\caption{Accuracy ($\%$) comparison on the BP4D-Spontaneous dataset.\vspace{0.1cm} \hspace{\textwidth} \hspace*{1.2cm}(a) Recognition \hspace{1cm} (b) Cross-Dataset Evaluation\vspace{-0.15cm}}
	\label{table:4DFERresults_part2}
	\resizebox{\linewidth}{!}{%
		\begin{tabular}{l c c}
			\hline
			Method & Accuracy (\color{blue}$\uparrow$\color{red}$\downarrow$\color{black})\\
			\hline
			{\color{teal}Yao \etal \cite{10.1145/3131345}} & 86.59 (\color{blue}1.75$\uparrow$\color{black})\\ 
			{\color{teal}Danelakis \etal \cite{danelakis2016effective}} & 88.56 (\color{red}0.22$\downarrow$\color{black})\\ 
			\hline
            \color{magenta}MiFaR \cite{behzad2021self} & 87.14 ({\color{blue}1.20$\uparrow$})\\ 
            \textbf{{\color{magenta}MultiviewVLM (Ours)}} & \textbf{88.34}\\ 
			\hline
		\end{tabular}
		\begin{tabular}{l c c}
			\hline
			Method & Accuracy (\color{blue}$\uparrow$\color{red}$\downarrow$\color{black})\\
			\hline
			{\color{teal}Zhang \etal \cite{ZHANG2014692}} & 71.00 (\color{blue}9.36$\uparrow$\color{black})\\ 
			{\color{teal}Zhen \etal \cite{zhen2017magnifying}} & 81.70 (\color{red}1.34$\downarrow$\color{black})\\ 
			\hline
            \color{magenta}MiFaR \cite{behzad2021self} & 79.05 ({\color{blue}1.31$\uparrow$})\\ 
            \textbf{{\color{magenta}MultiviewVLM (Ours)}} & \textbf{80.36}\\ 
			\hline
		\end{tabular}
	}
\end{table}
\subsection{Towards Spontaneous 4D FER}
To validate our model's capability in recognizing spontaneous expressions, we perform experiments on the BP4D Spontaneous dataset. This dataset consists of 41 subjects exhibiting spontaneous facial expressions, including two additional emotion classes: nervousness and pain. In this regard, Table~\ref{table:4DFERresults_part2} presents our extensive recognition and cross-dataset evaluation results. For the recognition task, our proposed MultiviewVLM model achieves the highest accuracy of 88.34\%, outperforming the previous unsupervised method MiFaR \cite{behzad2021self} (87.14\%) by \color{blue}1.20\%\color{black}, and improving over the supervised state-of-the-art method \cite{10.1145/3131345} (86.59\%) by \color{blue}1.75\%\color{black}. Although our model trails slightly behind \cite{danelakis2016effective} (88.56\%) by \color{red}0.22\%\color{black}, it still demonstrates competing performance among unsupervised methods, further advocating the strength of our multiview learning approach in capturing spontaneous facial dynamics.

Beyond recognition, we assess the generalizability of our model through a cross-dataset evaluation setting. Following the protocol of prior studies \cite{ZHANG2014692, zhen2017magnifying}, we train our model on the BU-4DFE dataset and validate it on selected tasks (Task 1 and Task 8) from BP4D-Spontaneous dataset, which include the happy and disgust emotions. This setup tests the model’s adaptability across different datasets with varied emotional contexts and subject populations, ultimately making the comparison more robust. Our model achieves an accuracy of 80.36\%, exceeding the competing models \cite{ZHANG2014692} (71.00\%) by \color{blue}9.36\% \color{black} and improving over the unsupervised baseline MiFaR~\cite{behzad2021self} (79.05\%) by \color{blue}1.31\%\color{black}. This is worth mentioning that while MultiviewVLM performs slightly below the state-of-the-art supervised method \cite{zhen2017magnifying} (81.70\%) by only \color{blue}1.34\%\color{black}, our results are overall still dominating given that our approach does not rely on manually annotated or labeled data.
\begin{figure}[t!]
    \centering
	\includegraphics[width=\linewidth]{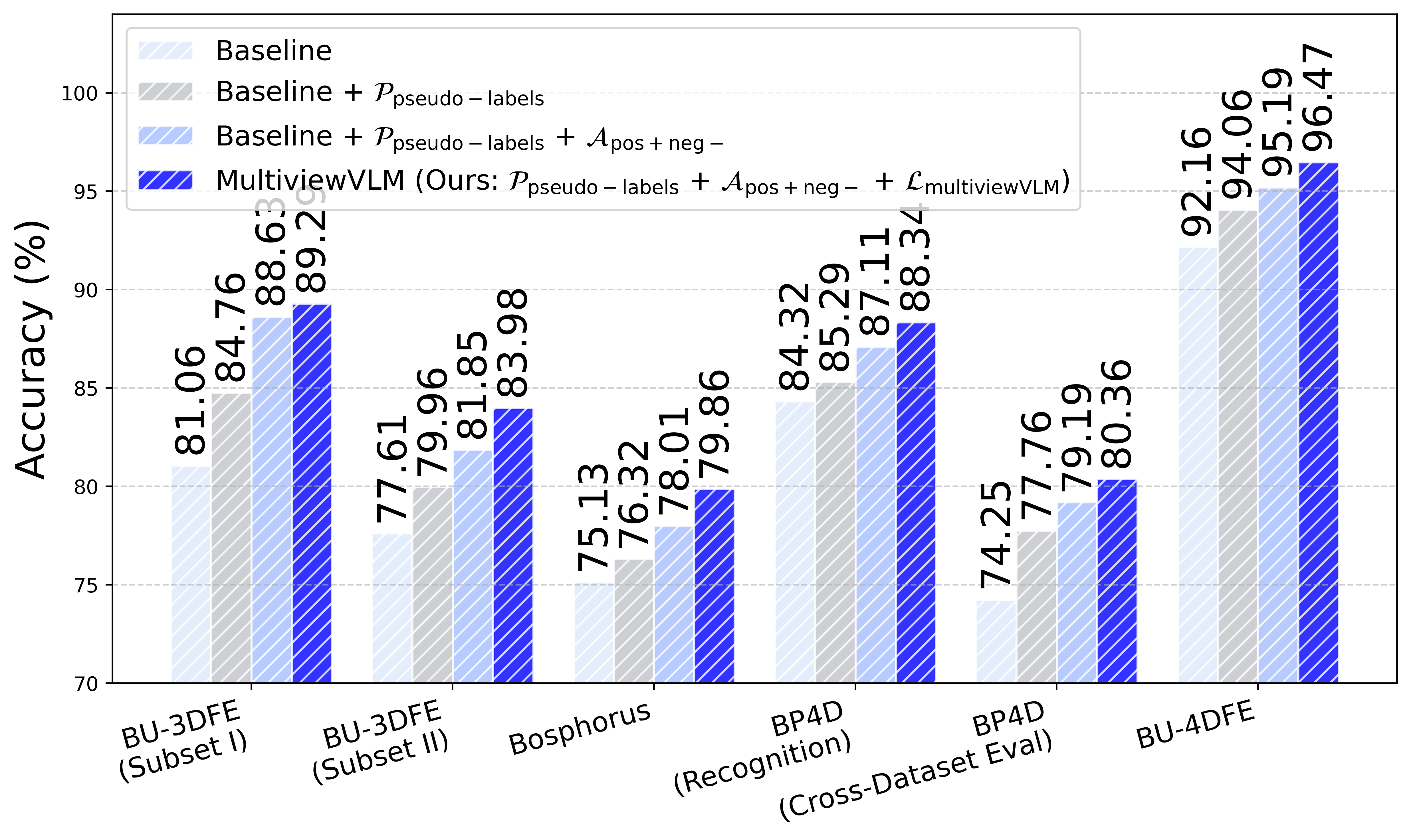}
	\caption{Ablation comparison of each proposed component of MultiviewVLM.}
	\label{fig:ablation}
\end{figure}

These findings collectively demonstrate that our model not only excels in recognizing spontaneous expressions but also generalizes effectively across different domains. The strong performance in both recognition and cross-dataset evaluation tasks highlights the robustness and transferability of the learned multiview representations from both visual and textual semantics. This positions MultiviewVLM as a highly promising unsupervised alternative for real world FER applications, especially where spontaneous expressions and limited annotations are common challenges that significantly impact model development.

\begin{figure}[b!]
    \centering
	\includegraphics[width=\linewidth]{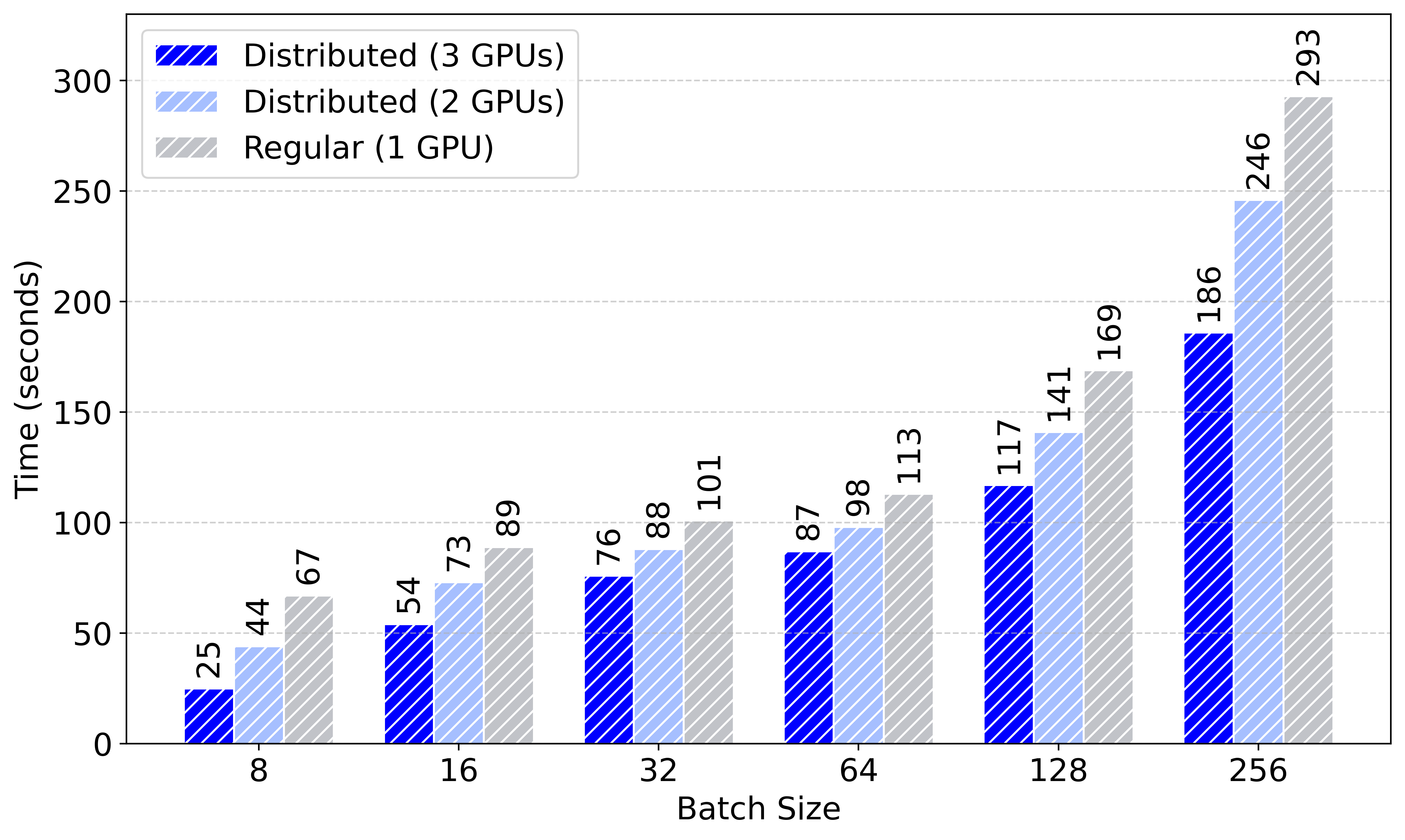}
	\caption{Performance comparisons of distributed learning.}
	\label{speedbar}
\end{figure}

\subsection{Ablation Study}
\textbf{Effectiveness of Each Component in MultiviewVLM:} To evaluate the contribution of each individual component in our proposed MultiviewVLM framework, we conduct a detailed ablation study across six benchmark datasets. As illustrated in Fig.~\ref{fig:ablation}, the results show a clear and consistent performance gain as each components is added. From a baseline model, the inclusion of pseudo-labels \( \mathcal{P}_{\mathrm{pseudo\text{-}labels}} \) provides a noticeable improvement in accuracy across all datasets. Furthermore, adding the positive-negative alignment module \( \mathcal{A}_{\mathrm{pos+neg-}} \) further enhances effective learning, especially evident in BP4D and BU-3DFE subsets. Finally, with our full MultiviewVLM model integrating pseudo-labels, positive-negative alignment, and the proposed multiview contrastive loss \( \mathcal{L}_{\mathrm{multiviewVLM}} \), the model achieves the highest performance in all settings, with the most pronounced gains. This demonstrates the significance of each component and the effectiveness achieved when they are combined.

\begin{figure*}[h!]
    \centering
	\includegraphics[width=\linewidth]{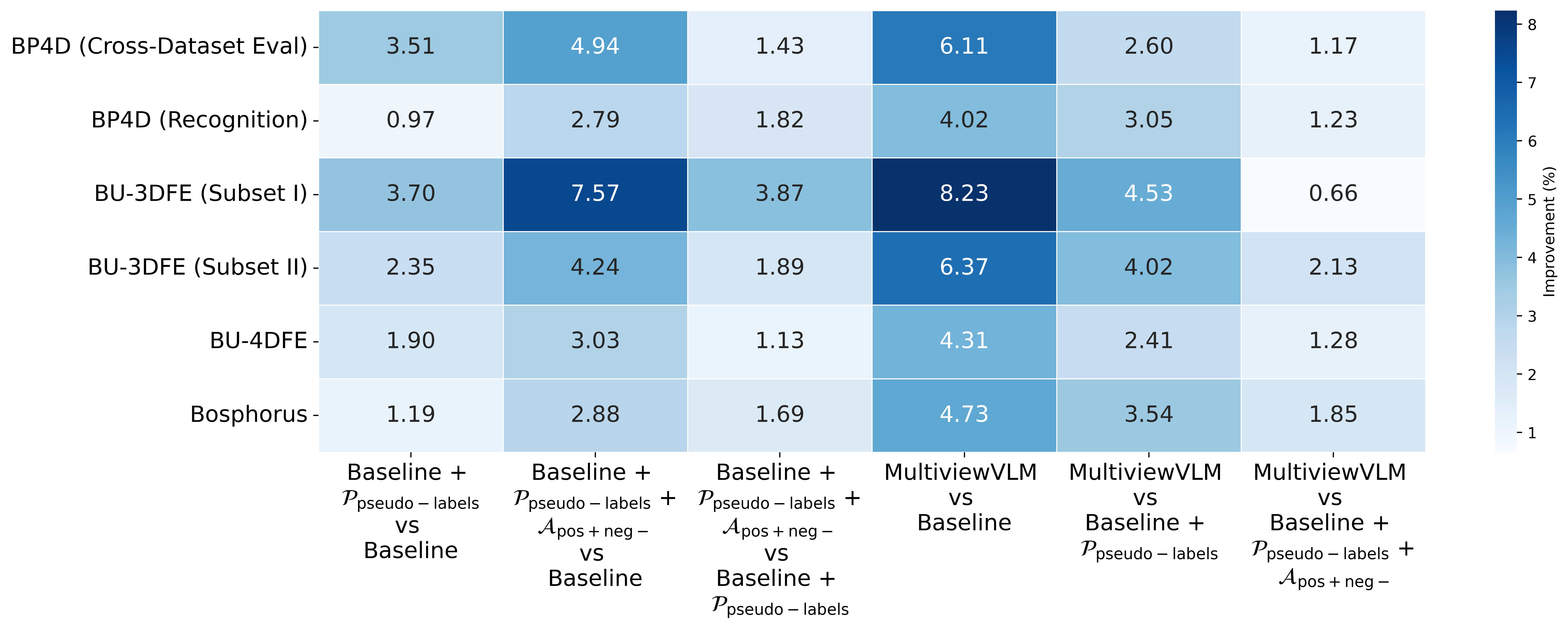}
	\caption{Accuracy improvement heatmap (bluish tone: higher = more improvement).}
	\label{accuracy_imp}
\end{figure*}

\textbf{Accuracy Improvements Across Datasets:} In Fig.~\ref{accuracy_imp}, we present a heatmap comparing accuracy improvements obtained by enhancing a baseline model using pseudo-labels $\mathcal{P}_{\mathrm{pseudo-labels}}$, positive-negative alignment $\mathcal{A}_{\mathrm{pos+neg-}}$, and the proposed MultiviewVLM framework containing all the ablation components. The heatmap uses a blue gradient to visually represent improvement levels, where darker tones signify greater accuracy gains. As shown in this figure, the most notable gains are observed on the BU-3DFE (Subset I) dataset, where the our proposed MultiviewVLM model yields over an 8\% improvement. On other datasets, such as BP4D (Recognition), the model shows moderate but consistent gains demonstrating the effectiveness of the our enhancements. This illustration confirms that the combination of joint unsupervised multiview learning offers substantial benefits across diverse 3D/4D facial expression datasets.
\begin{figure}[b!]
    \centering
	\includegraphics[width=\linewidth]{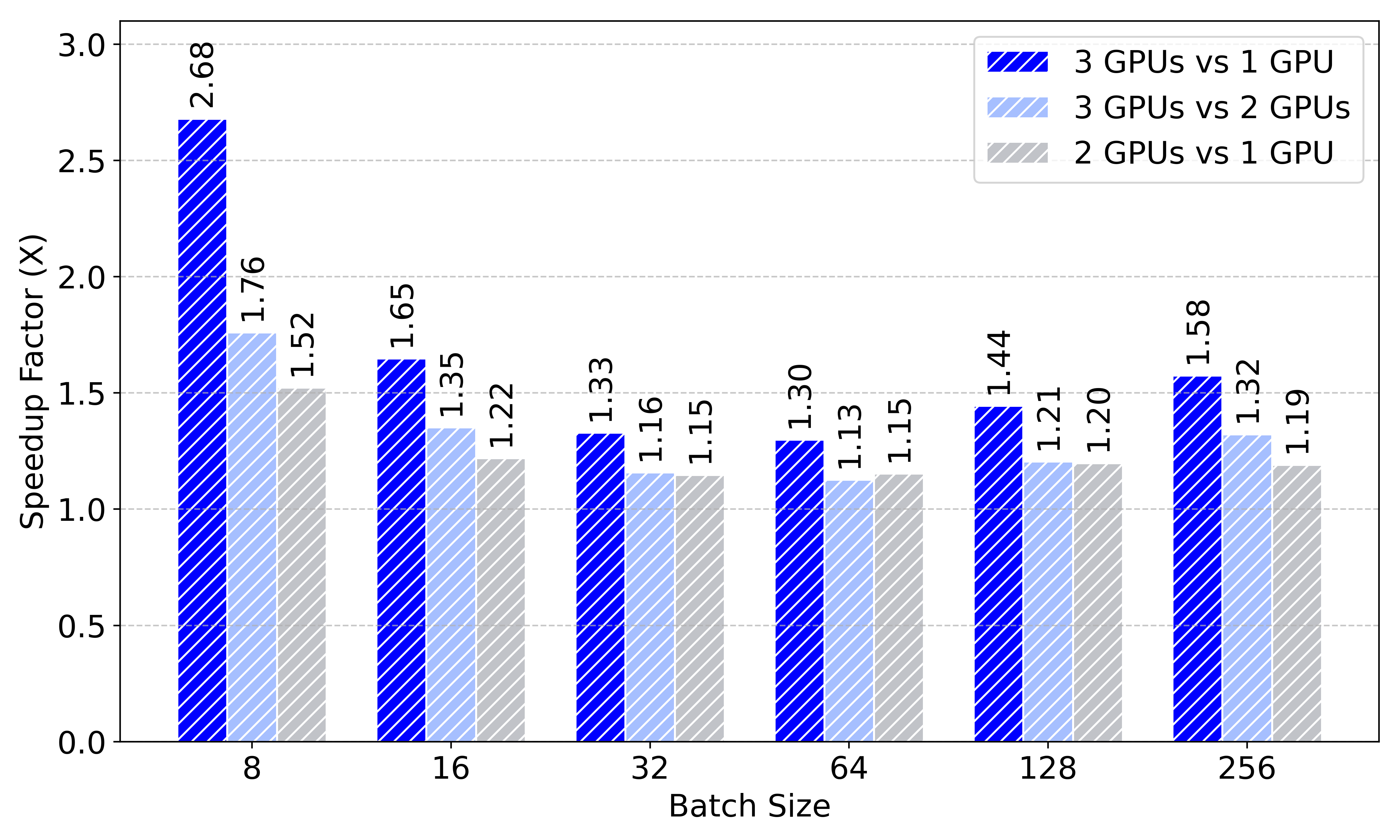}
	\caption{Speedup gains and comparison across various GPU configurations.}
	\label{gpuspeedbar}
\end{figure}

\textbf{Performance Upgrade with Distributed Learning:} To assess the computational efficiency of our model in realistic settings, we conducted training experiments using NVIDIA GeForce RTX 3090 Ti GPUs, which are relatively powerful but are not considered ultra high-end compared to more recent GPU architectures. This choice is intentional in order to demonstrate that our model can achieve strong performance without requiring top-tier hardware. To examine the performance under these constrained training conditions, we evaluate the training time across three hardware configurations: distributed training with 3 GPUs, distributed training with 2 GPUs, and a regular single-GPU training across varying batch sizes of 8, 16, 32, 64, 128 and 256. As shown in Fig.~\ref{speedbar}, for a batch size of 8, the 3-GPU distributed setup completes an epoch in just 25 seconds, compared to 44 seconds on 2 GPUs and 67 seconds on a single GPU. This marks a 168\% speedup over the regular training configuration. At a batch size of 16, the 3-GPU configuration again leads with 54 seconds per epoch, while the 2-GPU and 1-GPU setups take 73 and 89 seconds, respectively, reflecting consistent gains of 35\% and 65\%. Similarly, for a batch size of 32, training takes 76 seconds on 3 GPUs, compared to 88 seconds with 2 GPUs and 101 seconds on a single GPU. As further illustrated in Fig.~\ref{speedbar} for the batch size of 64, 128, and 256, we show that our distributed 3-GPU setup consistently outperforms other provided setups. These results reinforce the effectiveness of our distributed training strategy, showing that performance benefits persist even at small batch sizes where communication overhead is typically more noticeable. The clear trend across all three configurations confirms that our model remains highly scalable and efficient, regardless of batch size constraints.

When extrapolated over 500 training epochs, the efficiency gains become even more pronounced. For instance, with a batch size of 64, the distributed training with 3 GPUs takes approximately 15.7 hours, compared to over 40.7 hours on a single GPU. Similarly, with batch sizes of 128 and 256, the total training time drops to 13.6 hours and 12.1 hours, respectively, with the 3-GPU setup, while the single-GPU training take 34.2 hours and 25.8 hours. These results validate the scalability and robustness of our model under distributed training making it accessible and practical for deployment even in moderate hardware environments. The substantial reductions in training time also provide accelerated research directions for faster experimentation cycles and greater adaptability in real world applications where computational resources may be limited especially on unannotated data.

\textbf{Speedup gains with various GPU configurations:}
In Fig.~\ref{gpuspeedbar}, we illustrate the speedup factors achieved across various batch sizes when training with different GPU configurations. Specifically, the performance of distributed training using 3 GPUs is compared against 2 GPUs and a single GPU settings. At smaller batch sizes, the speedup gains from using 3 GPUs over 1 GPU is most significant, reaching up to 2.68× and 1.65× for batch size 8 and 16, respectively. We also notice that as the batch size increases, the relative gains slightly stabilize around the range of 1.3×–1.6×. Similarly, 3 GPUs outperform 2 GPUs consistently with speedup gains ranging from 1.13× to 1.76×. The comparison between 2 GPUs and 1 GPU also follows a similar trend, with improvements between 1.15× and 1.52×. These results indicate that while distributed training offers the greatest benefits at lower batch sizes due to reduced per-epoch time, it continues to provide stable and meaningful improvements across all configurations, validating its efficiency and scalability for diverse workload intensities.

\textbf{Towards Explainability with Grad-CAM Visualizations:} Towards explainability of our model, we use Grad-CAM~\cite{selvaraju2017grad} visualizations that showcase the attention patterns learned by our proposed model for 3D/4D FER on multiview data. We train our model on the BP4D-Spontaneous dataset and perform zero-shot inference on the BU-4DFE dataset. The visualizations are randomly scattered across various subjects performing one of the six prototypical facial expressions captured from three distinct angles: front, left, and right profiles. The Grad-CAM method helps to extract and visualize the most discriminative regions of each image that influence the model's predictions.
\begin{figure*}[t!]
    \centering
	\includegraphics[width=\linewidth]{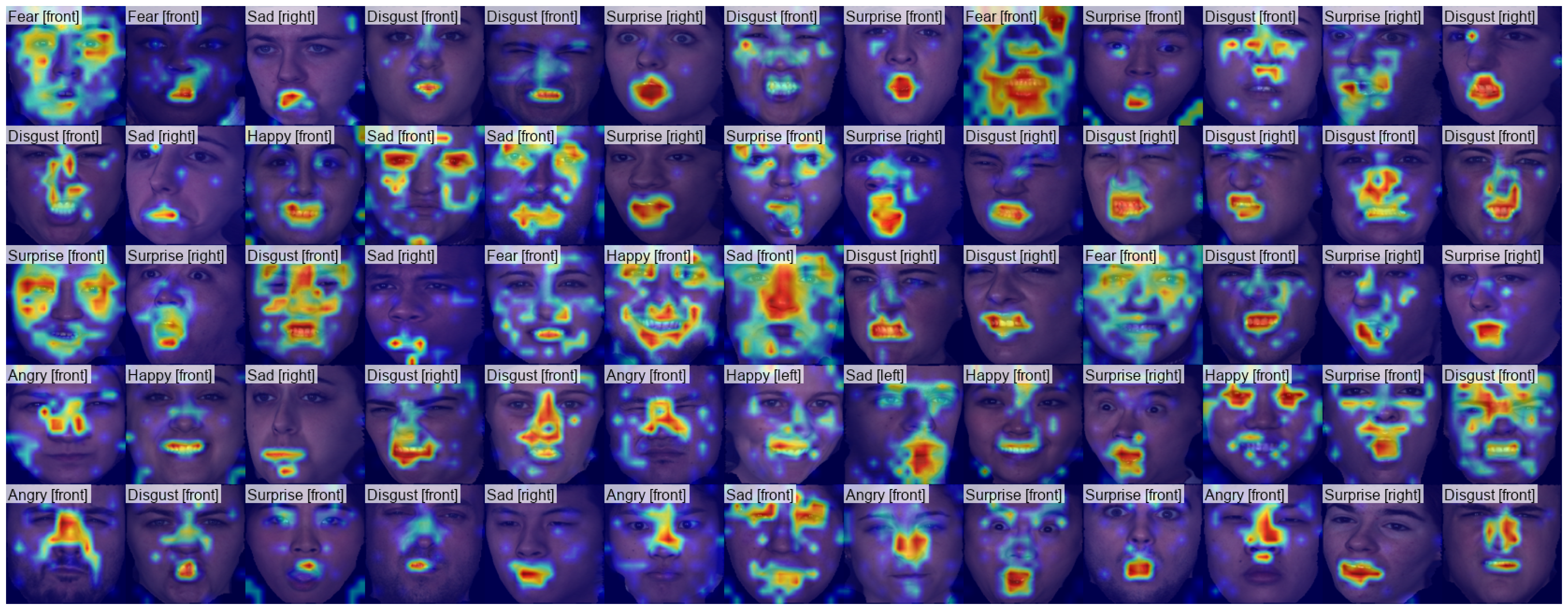}
	\caption{Grad-CAM visualization of six facial expressions (Happy, Sad, Angry, Fear, Surprise, Disgust) across three viewpoints (front, left, right). Each row contains 13 randomly selected images, with five total rows covering a diverse mix of identities, emotions, and perspectives. The heatmaps highlight the spatial attention of our proposed model for each input image, showcasing where the model focuses to make predictions. The overlaid labels specify both the ground-truth emotion and the corresponding viewpoint of the subject.}
	\label{gradcam_vis}
\end{figure*}

From the heatmaps shown in Fig.~\ref{gradcam_vis}, it is evident that the model primarily focuses on the eyes, eyebrows, nose, and mouth regions, which are the most informative for distinguishing among facial expressions. For instance, the emotion ``Surprise" is consistently associated with high attention around the widened eyes and open mouth, while the emotion ``Disgust" highlights activations near the nose and upper lip. Similarly, the ``Sad" emotion class tends to emphasize the bending eyes and corners of the mouth, while the emotion ``Anger" focuses usually on the wrinkled forehead and tense lips. This spatialy-captured behavior aligns very well with psychological theories of facial expression encoding (e.g., FACS) \cite{li2018deep}. 

Interestingly, the visualizations also demonstrate the ability of our proposed MultiviewVLM model to maintain consistent attention across different viewpoints. For instance, while frontal views offer the richest feature details, left and right profiles still preserve enough muscle geometry for accurate emotion detection. The consistency of high-activation zones across different views suggests that our model has learned view-invariant representations of emotional cues. This behavior validates the robustness of the learned feature space and the effectiveness of our novel contributions for multiview recognition models.

\section{Conclusion}
We presented MultiviewVLM, a novel vision-language model architecture that leverages multiview data in an unsupervised setting for enhanced understanding of 3D/4D facial expression recognition. We introduced pseudo-labels which are derived from generated text prompts ultimately enabling implicit semantic alignment of emotion representations. Supported by our multiview contrastive framework, we proposed a unified joint embedding space representation that integrates multiview features by learning from positive and negative pairs with numerical stability fostering discriminative and generalizable representations in an unsupervised setting. To further improve convergence, we introduced a gradient-friendly loss function, and designed the model for scalable distributed training. Comprehensive experimental results demonstrate that MultiviewVLM consistently outperforms existing state-of-the-art models across multiple benchmarks. Thanks to its modular and flexible design, the model can be readily adapted to a wide range of real-world applications with minimal customization efforts.

{\small
\bibliographystyle{ieeetr}
\bibliography{references}
}

\end{document}